\documentclass{article}
\usepackage{spconf,amsmath,epsfig}
\usepackage{hyperref}
\usepackage{amssymb}
\usepackage{amsmath}  
\usepackage{algorithm}
\usepackage{algpseudocode} 
\usepackage{adjustbox}
\usepackage{booktabs}
\usepackage{multirow}
\usepackage{url}
\usepackage{graphicx}
\usepackage{amsfonts}
\usepackage{booktabs, multirow, placeins, float, indentfirst, mathrsfs, tabularx, adjustbox}
\usepackage{array}
\usepackage{pifont}
\usepackage{cancel}
\usepackage{titlesec}
\usepackage{hyperref}

\let\OLDthebibliography\thebibliography
\renewcommand\thebibliography[1]{
  \OLDthebibliography{#1}
  \setlength{\parskip}{0pt}
  \setlength{\itemsep}{0pt plus 0.3ex}
}

\begin{document}\sloppy

% Title.
% ------
\title{Multi-modal Fusion and Query Refinement Network for Video Moment Retrieval and Highlight Detection}
% 记得最后改成匿名提交!!!!!!!!!!!!!!!!!!!!!!!!!!!!!!!!!!!
% \name{Anonymous ICME submission}
% \address{}
\name{
Yifang Xu$^{1,*}$, Yunzhuo Sun$^{2,*}$, Benxiang Zhai$^1$, Zien Xie$^1$, Youyao Jia$^{3}$, Sidan Du$^{1,\dag}$
\thanks{\textsuperscript{*}Equal contribution.}
\thanks{\textsuperscript{\dag}Corresponding author}
}
\address{
\small $^{1}$School of Electronic Science and Engineering, Nanjing University, China
\\
\small $^{2}$School of Physics and Electronics, Hubei Normal University, China
\\
\small $^{3}$Gosuncn Chuanglian Technology Co., Ltd., China
\\
\small \{xyf, xze, zbx\}@smail.nju.edu.cn; sunyunzhuo98@outlook.com; coff128@nju.edu.cn
}
\maketitle

% 100-150 words!

\begin{abstract}
Given a video and a linguistic query, video moment retrieval and highlight detection (MR\&HD) aim to locate all the relevant spans, while simultaneously predicting saliency scores. Most existing methods utilize RGB images as input, overlooking the inherent multi-modal visual signals like optical flow and depth. In this paper, we propose a \textbf{M}ulti-modal Fusion and Query \textbf{R}efinement \textbf{Net}work (\textbf{MRNet}) to learn complementary information from multi-modal cues. Specifically, we design a multi-modal fusion module to dynamically combine RGB, optical flow, and depth map. Furthermore, to simulate human understanding of sentences, we introduce a query refinement module that merges text at different granularities, containing word-, phrase-, and sentence-wise levels. Comprehensive experiments on QVHighlights and Charades datasets indicate that MRNet outperforms current SOTA methods, achieving notable improvements in MR-mAP@Avg (\textbf{+3.41}) and HD-HIT@1 (\textbf{+3.46}) on QVHighlights. 
% Our code will be available at -. 

\end{abstract}

\vspace{-1mm}
\section{Introduction}
\label{sec:intro}
\vspace{-2mm}

\begin{figure}[t!]
  \centering
  \includegraphics[width=\linewidth]{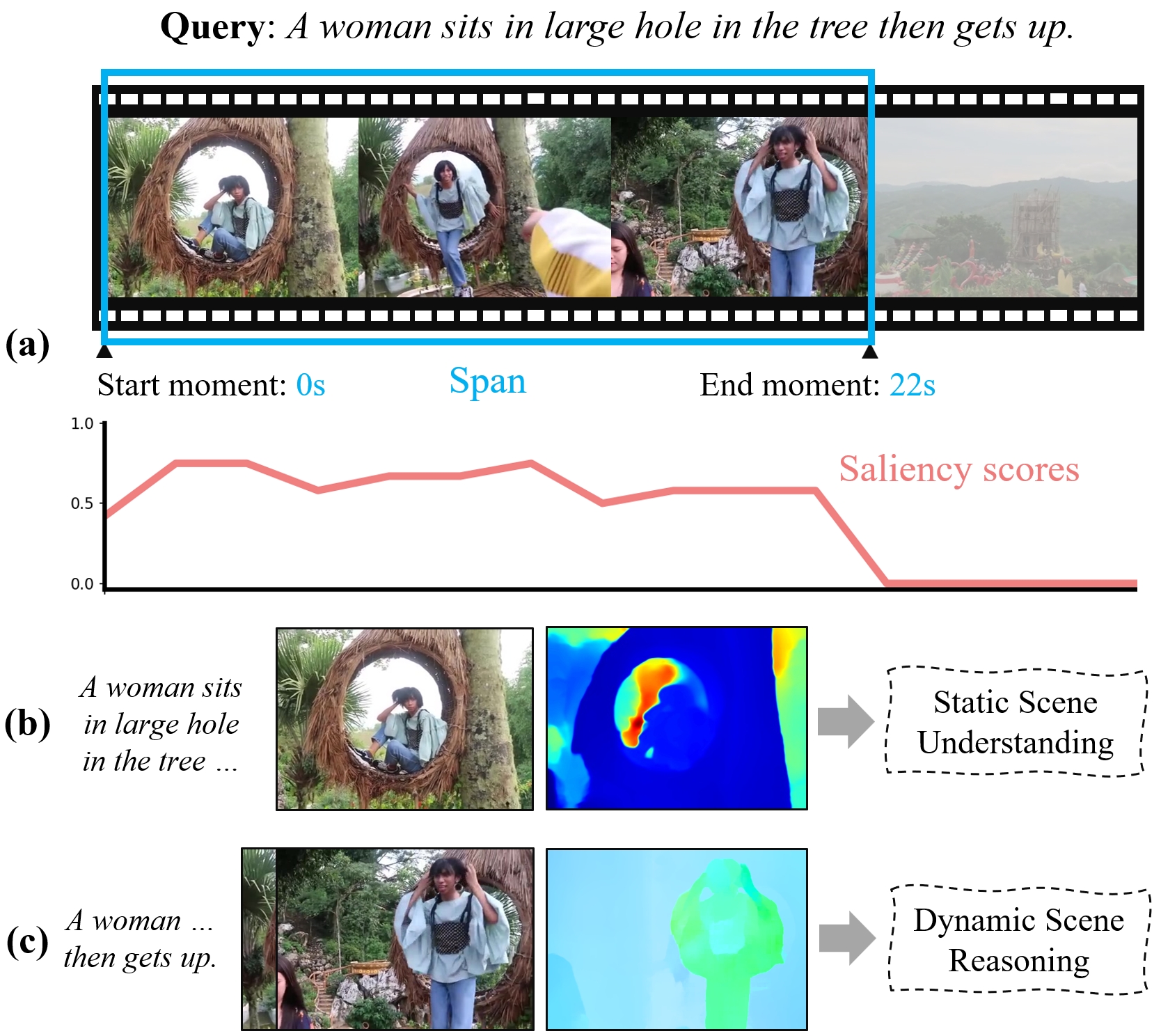}
  \vspace{-6mm}
  \caption{
    (a) An depiction of MR\&HD. (b) Depth information enhances the model to understand static scene. (c) Optical flow reinforces the model to reason about dynamic scene.
  }
  \label{Fig:MRHD}
  \vspace{-3mm}
\end{figure}

Users on video platforms sift through countless videos daily and are eager to retrieve relevant highlight frames based on textual query, facilitating their efficient video browsing. In this paper, we focus on two sub-tasks in video understanding: moment retrieval (MR) and highlight detection (HD). Given a textual query, MR aims to locate all the relevant temporal spans from a video, where each span contains a start and end moment \cite{MCN-2017}. HD aims to obtain saliency scores for each frame \cite{LIM-S-2019}. An illustrative example of MR\&HD is shown in Fig. \ref{Fig:MRHD} (a).

\begin{figure*}[t!]
  \centering
  \includegraphics[width=\linewidth]{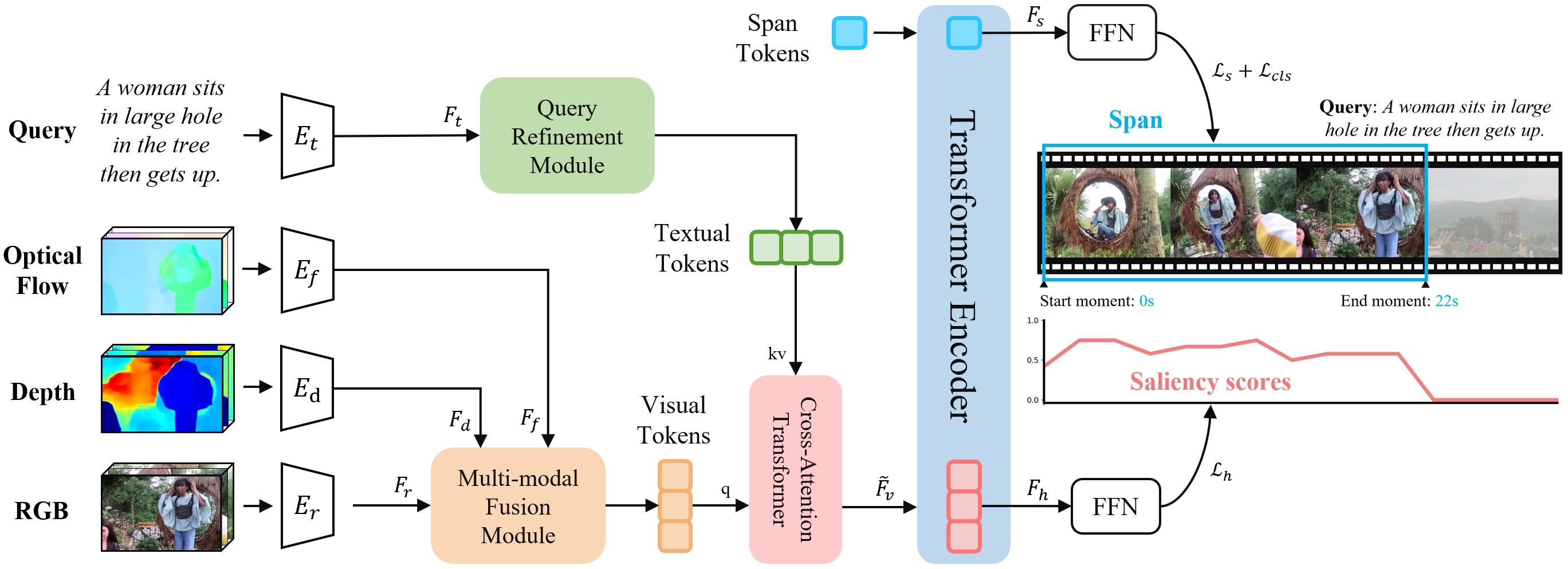}
  \vspace{-5mm}
  \caption{
    % 123
    Overview of Multi-modal Fusion and Query Refinement Network (MRNet). 
    % 这个网络由
    % For input video, we first compute the depth map for each frame and the optical flow between consecutive frames. Subsequently, we employ visual extractors ($E_r$, $E_d$, $E_f$) to derive corresponding features $F_r$, $F_d$, and $F_f$ from the RGB, depth, and optical flow inputs, respectively. The multi-modal fusion module then fuses these features to obtain visual tokens. For input query text, we utilize the textual extractor $E_t$ to get word-wise features $F_t$. The query refinement module then generates phrase- and sentence-wise features from $F_t$ and aggregates them to produce textual tokens $\tilde{F}_t$. Both visual and textual tokens are fed into cross-attention transformer to get query-relevant cross-modal tokens $\tilde{F}_v$. Finally, we apply transformer encoder and prediction heads to obtain results.
  }
  \label{Fig:MRNet}
  \vspace{-3mm}
\end{figure*}

Existing methods \cite{MomentDETR-2021, UMT-2022, MomentDiff-2023} typically only use RGB images as visual features and overlook the naturally existing multi-modal cues in videos, which leads to poor model performance in videos containing rich moving objects and cluttered backgrounds. Compared to RGB, optical flow provides motion information, which enhances the model to reason about dynamic scenes in videos \cite{GMFlow-2022}
% \cite{RAFT-2022, GMFlow-2022}. 
As depicted in Fig. \ref{Fig:MRHD} (c), dynamic scenes \textit{"A woman ... then gets up"} involves human motion, and utilizing optical flow enables the model to infer such dynamic behavior. Additionally, RGB images may be affected by lighting and shadows, while depth maps are relatively more reliable. Depth maps contain objects distance within a scene, which is beneficial for the model to recognize object movements with distinct shapes in video understanding tasks \cite{DCTNet-2022, PFANet-2021}. As shown in Fig. \ref{Fig:MRHD} (b), extracting optical flow from static scenes with cluttered backgrounds such as \textit{"A woman sits in large hole in the tree ..."} is challenging. However, the spatial structural information provided by depth maps can assist the model in understanding static scenes.

% "In order to extract comprehensive insights from video data, MRNet adeptly incorporates a range of visual inputs, encompassing RGB imagery, optical flow, and depth perception."

In this paper, we newly propose the \textbf{M}ulti-modal Fusion and Query \textbf{R}efinement \textbf{Net}work (\textbf{MRNet}) to address the above challenges in MR\&HD. To learn complementary information from video, MRNet integrates various visual inputs, including RGB, optical flow, and depth. Furthermore, we introduce a query refinement module inspired by the layered way in which humans understand language \cite{frankHowHierarchicalLanguage2012, miyagawaEmergenceHierarchicalStructure2013}. It extracts word-, phrase-, and sentence-wise features from linguistic queries to exploit semantic information under a global receptive field. Overall, our main contributions are as follows:

1) We present an MR\&HD model, MRNet, which fuses RGB, optical flow, and depth features to reinforce dynamic scene reasoning and static scene understanding.

2) We design a query refinement module to exploit textual features at different levels, including words, phrases, and sentences.

3) Comprehensive experiments on QVHighlights and Charades dataset indicate that MRNet outperforms existing SOTA methods.

\vspace{-3mm}
\section{Related work}
\vspace{-3mm}

Most previous MR\&HD  approaches \cite{MomentDiff-2023, sun2023gptsee, MH-DETR-2023} only employ image and text inputs. To mine multi-modal information, UMT \cite{UMT-2022} exploits audio signal and designs a more unified architecture for MR\&HD. Some recent works \cite{DCTNet-2022, Study-Multi-modal-2023} reveal that leveraging multiple visual cues can help the model understand videos. DCTNet \cite{DCTNet-2022} harnesses flow and depth to enrich spatio-temporal context in video salient object detection. VIOLETv2 \cite{Study-Multi-modal-2023} integrates flow and depth in visual pre-training to enhance performance on various downstream tasks, ranging from video question answering to video captioning, et al.

DETR-based methods \cite{DETR, MomentDETR-2021, QRNet-2022}, commonly utilized for localization tasks like object detection (OD), visual grounding (VG), and MR, have been observed to face slow convergence issues. To address this limitation, TSP \cite{TSP-2021} introduces a transformer encoder-only architecture to expedite the training convergence of DETR in OD. QRNet \cite{QRNet-2022} develops a decoder-free visual-linguistic transformer tailored for VG to extract query-consistent visual features.

\vspace{-2mm}
\section{Method}
\vspace{-2mm}

% In this section, we first formulate the HD and MR tasks and then present the overall architecture of our model. Subsequently, we provided details of each module in the model. Finally, we briefly introduce the training loss.

\subsection{Overview}
\vspace{-2mm}
For a video $V \in \mathbb{R}^{N_{v} \times H \times W \times 3}$ containing $N_v$ moments and a corresponding query $T \in \mathbb{R}^{N_{t}}$ containing $N_{t}$ words. MR\&HD aims to retrieve all video spans $S \in \mathbb{R}^{N_{s} \times 2}$ that are highly relevant to $T$ (each span comprises a starting and ending moment), while concurrently computing moment-wise saliency scores $H \in \mathbb{R}^{N_{v}}$.  As shown in Fig. \ref{Fig:MRNet}, our proposed MRNet can be divided into five components: multi-modal fusion module, query refinement module, cross-attention transformer, transformer encoder, and prediction heads. 

For input videos, we first use ZoeDepth \cite{ZoeDepth-2023} to predict the depth map of each frame and utilize GMFlow \cite{GMFlow-2022} to compute the optical flow between each pair of adjacent frames. Then, we use the frozen CLIP \cite{CLIP-2021} video encoder ($E_{r}$, $E_{d}$, $E_{f}$) to extract features $F_{r}$, $F_{d}$, and $F_{f} \in \mathbb{R}^{N_{v} \times d}$ from the RGB, depth, and optical flow, respectively. For input queries, we employ the text encoder of CLIP $E_{t}$ to derive word-wise features $F_{t} \in \mathbb{R}^{N_{t} \times d}$. Subsequently, the multi-modal fusion module merges $F_{r}$, $F_{d}$, and $F_{f}$ to generate visual tokens $F_{v} \in \mathbb{R}^{N_{v} \times d}$. The query refinement module generates word-, phrase-, and sentence-wise features from $F_{t}$ and aggregates them to obtain textual tokens $\tilde{F}_{t} \in \mathbb{R}^{(N_{t}+1) \times d}$. The visual and textual tokens are fed to cross-attention transformer to generate query-relevant cross-modal tokens $\tilde{F}_{v} \in \mathbb{R}^{N_{v} \times d}$. Finally, we apply transformer and prediction heads to obtain MR and HD results.

\begin{figure}[t!]
  \centering
  \includegraphics[width=0.8\linewidth]{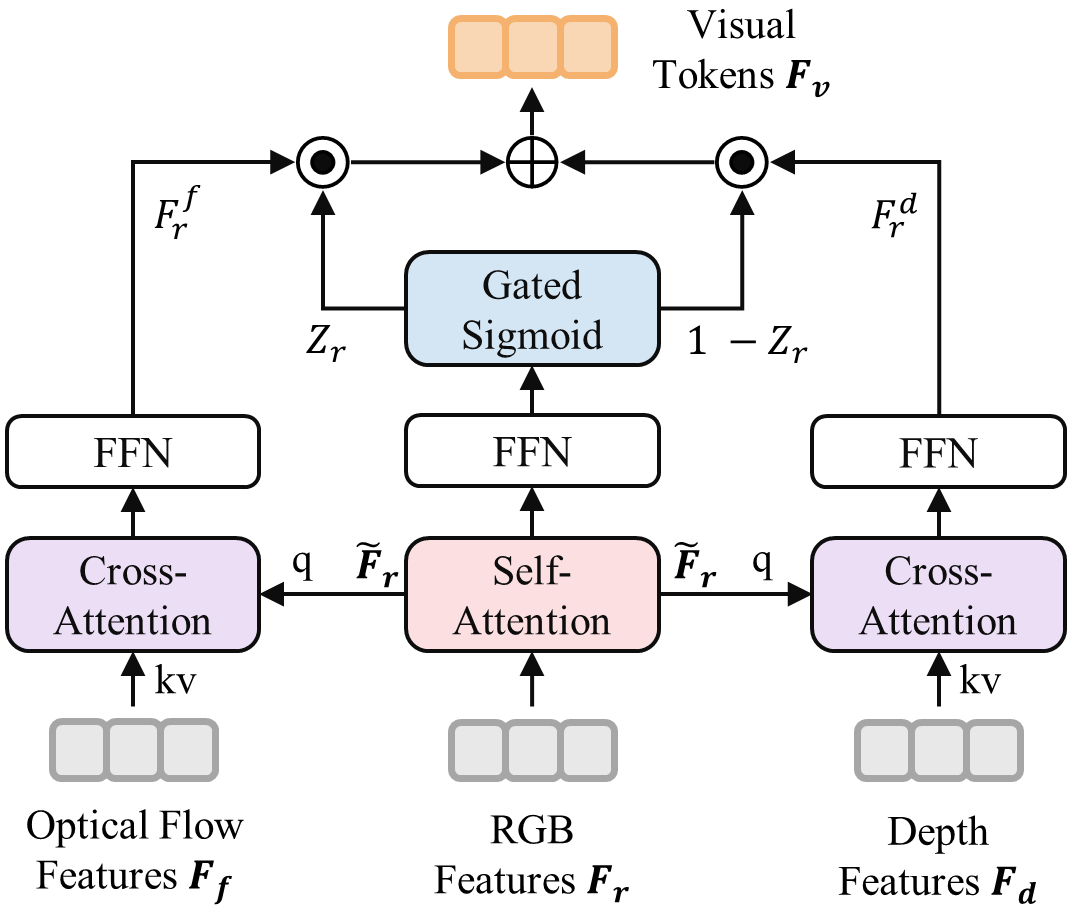}
  \vspace{-2mm}
  \caption{
    The multi-modal fusion module (MFM) aggregates RGB, optical flow, and depth features to enhance dynamic scene reasoning and improve static scene understanding.
  }
  \label{Fig:MFM}
  \vspace{-2mm}
\end{figure}

\vspace{-2mm}
\subsection{Multi-modal Fusion Module}
\label{sec:MFM}
\vspace{-1mm}

Videos inherently contain rich multi-modal cues that provide valuable information. Compared to RGB images, optical flow and depth maps, which respectively provide information on object motion and scene structure, can improve dynamic scene inference and static scene understanding, as demonstrated in Fig. \ref{Fig:MRHD}. Most existing works use CNN \cite{RASL-2019} or co-attention \cite{DCTNet-2022} with late fusion to aggregate these three modalities. However, in our experiments, we observed that co-attention with late fusion does not yield satisfactory results, as shown in Tab. \ref{tab:ablation-visual-encoder}. Furthermore, models that only use flow or depth features as input also have mediocre performance, as described in Tab. \ref{tab:ablation-features}. This may be due to the fact that flow and depth themselves cannot capture semantic information and are therefore not suitable as semantic queries. Recent work \cite{Study-Multi-modal-2023} also indicates that directly concatenating RGB, optical flow, and depth as inputs to transformer does not necessarily improve the performance of video-language models.

To tackle the above problems, we design a multi-modal fusion module (MFM) that dynamically fuses RGB, flow, and depth features, as depicted in Fig. \ref{Fig:MFM}. Firstly, given that RGB features generally encompass the most abundant information in video, we incorporate self-attention to model the video context with a global receptive field. Subsequently, we utilize contextual RGB features $\tilde{F}_{r}$ for cross-attention \textit{query}, and flow features $F_{f}$ or depth features $F_{d}$ as \textit{key} and \textit{value}, respectively. This enables the generation of the RGB-guided flow features $F_{r}^{f} \in \mathbb{R}^{N_{v} \times d}$ and depth features $F_{r}^{d} \in \mathbb{R}^{N_{v} \times d}$, thereby mitigating the performance degradation caused by the lack of semantic information in flow and depth features. 
% Following previous works \cite{GRU-2014, MIGCN-2021}, 
We finally employ a gated sigmoid $Z_{r}$ to dynamically balance $F_{r}^{f}$ and $F_{r}^{d}$. The final visual tokens $F_{v}$ can be formulated as: 
\begin{equation}
    F_{v} = Z_{r} \odot F_{r}^{f} + (1 - Z_{r}) \odot F_{r}^{d}
\end{equation}
where $Z_{r} = Sigmoid(FFN(\tilde{F}_{r}))$, $\odot$ denotes element-wise product. For brevity, we omit the residual structure, ReLU activation, and post-norm style layernorm \cite{LayerNorm-2016} in the attention layers and FFNs in both the figures and equations.

\begin{figure}[t!]
  \centering
  \includegraphics[width=\linewidth]{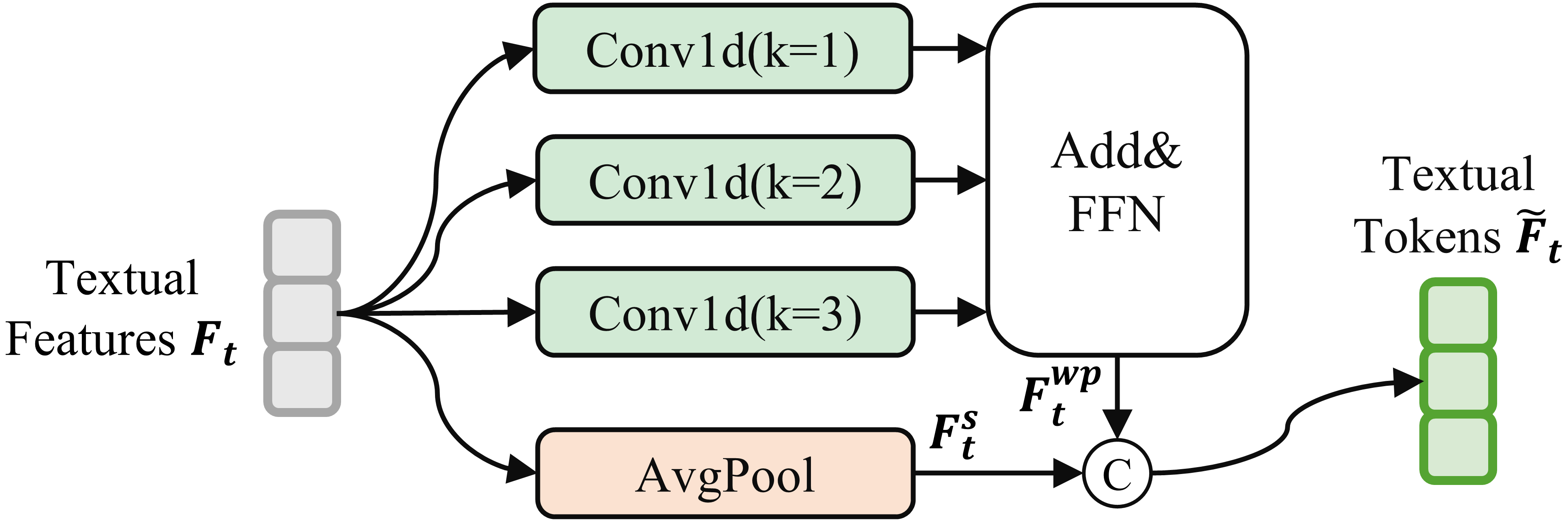}
  \vspace{-4mm}
  \caption{
    The query refinement module (QRM) integrates textual features at different levels.
  }
  \label{Fig:QRM}
  \vspace{-3mm}
\end{figure}

\vspace{-2mm}
\subsection{Query Refinement Module}
\vspace{-1mm}
MR\&HD bears a resemblance to the reading comprehension task in NLP, where predictions are highly correlated with natural language queries \cite{Survey-TSGV-2022}. The human reading habit usually understands the words and phrases while skimming the whole sentence. From a biological viewpoint, human language processing follows a layered hierarchy \cite{frankHowHierarchicalLanguage2012, miyagawaEmergenceHierarchicalStructure2013}. 

Building on the above works, we propose a query refinement module (QRM) to simulate human understanding of sentences, which fuses word-, phrase- and sentence-wise features from text. As illustrated in Fig \ref{Fig:QRM}, we employ a temporal 1D convolution to strengthen word-wise features. Additionally, to capture more semantic details, we also extract phrase-wise features. Since phrases typically consist of 2 to 3 words, we apply two 1D convolution operations with kernel sizes of 2 and 3. Next, we add the word- and phrase-wise features and feed them into FFN to produce aggregated features $F_{t}^{wp} \in \mathbb{R} ^ {N_{t} \times d}$. An average pooling operation is utilized to get sentence-wise features $F_{t}^{s}$. Finally, we concatenate $F_{t}^{wp}$ and $F_{t}^{s}$ to obtain textual tokens $\tilde{F}_{t} \in \mathbb{R} ^ {(N_{t} + 1)\times d}$ containing hierarchical information.

\vspace{-2mm}
\subsection{Cross-Attention Transformer and Transformer Encoder}
\vspace{-1mm}
To produce cross-modal tokens, Moment-DETR \cite{MomentDETR-2021} directly concatenates visual with textual tokens and feeds them into transformer. However, if the similarity between video moments greatly exceeds video-query similarity, the resulting tokens become unrelated to the query, compromising overall performance. To tackle this challenge, we apply a cross-attention transformer \cite{Transformer} (CAT) to generate query-relevant cross-modal tokens $\tilde{F}_{v} \in \mathbb{R}^{N_{v} \times d}$ containing highlighted and temporal information. Specifically, cross-attention layer computes attention weights between video moments and text tokens, allowing each moment to learn which concepts are present within text tokens. 

Subsequently, inspired by previous works \cite{TSP-2021, QRNet-2022, RSVG-2023}, 
% \cite{TSP-2021, QRNet-2022, RSVG-2023}, 
we remove the transformer decoder from Moment-DETR \cite{MomentDETR-2021} to create a decoder-free architecture for MR\&HD. In our experiments, we are surprised that this naive way achieves decent gain, as shown in Tab. \ref{tab:ablation-decoder}. More concretely, we concatenate $\tilde{F}_{v}$ with learnable span tokens $\tilde{S} \in \mathbb{R}^{N_{s} \times d}$ and feed them into transformer encoder \cite{Transformer}, which outputs highlight features $F_{h} \in \mathbb{R}^{N_{v} \times d}$ and span features $F_{s} \in \mathbb{R}^{N_{s} \times d}$. Next, we feed them into prediction heads to get the final results.

\vspace{-1mm}
\subsection{Prediction Heads and Training Loss}
\vspace{-1mm}
For highlight features $F_{h}$, we utilize 2-layer FFN to obtain saliency scores $H \in \mathbb{R}^{N_{v}}$. For span features $F_{s}$, we employ 2-layer FFN with sigmoid to predict the normalized spans $S \in \mathbb{R}^{N_{s} \times 2}$. A linear layer and softmax are applied for classification prediction. Moments consistent with the ground truth are labeled as \textit{foreground}, while the rest are denoted as \textit{background}.

The overall loss $\mathcal{L}$ combines highlight loss $\mathcal{L}_{h}$, span loss $\mathcal{L}_{s}$, and classification loss $N_{cls}$, modulated by hyperparameters $\lambda_{*}$:
\begin{equation}
    \mathcal{L} = \lambda_{h} \mathcal{L}_{h} + \mathcal{L}_{s} + \lambda_{cls} \mathcal{L}_{cls}
\end{equation}
For the \textit{i}-th moment, we define its predicted saliency score and ground truth as $\hat{h}_{i}$ and $h_i \in [ 0, 1 ]$, respectively. The highlight loss $\mathcal{L}_{h}$ as follows:
\begin{equation}
    \mathcal{L}_{h} = -\sum_{i=1}^{N_{v}} \left[ h_i \log (\hat{h}_{i}) + \left( 1-h_i \right) \log (1-\hat{h}_{i}) \right]
\end{equation}
Leveraging the Hungarian algorithm as in \cite{DETR}, we match predicted spans $\hat{s}$ with ground-truth  $s$. For the $N_{m}$ matched pairs within a video, $\mathcal{L}_{s}$ measures discrepancies using both L1 loss and IoU loss. 
% \cite{GIoU-2019}. 
Additionally, we apply $\mathcal{L}_{cls}$ to classify the predicted span to \textit{foreground} or \textit{background}:
\begin{equation}
    \mathcal{L}_{s} = \sum_{i=1}^{N_{m}} \left[ \lambda_{L1}|| \hat{s} - s ||_{1} + \lambda_{IoU} \mathcal{L}_{IoU}(\hat{s}, s) \right]
\end{equation}
\begin{equation}
    \mathcal{L}_{cls} = - \sum_{i=1}^{N_{s}}  \left[ w_{p} y_{i} \log(p_{i}) + (1 - y_{i}) \log(1 - p_{i}) \right]
\end{equation}
Here, $p_{i}$ denotes \textit{foreground} probability, and $y_{i} \in \{ 0, 1 \}$ is its label. To address label imbalance, \textit{foreground} is weighted more by $w_{p}$.

\vspace{-1mm}
\section{Experiments}
\vspace{-1mm}

\subsection{Dataset and Evaluation Metrics}
\vspace{-1mm}
\textbf{QVHighlights} \cite{MomentDETR-2021} is currently the only dataset supporting both HD and MR tasks. It comprises about 10,000 distinct YouTube videos, all annotated with textual queries, saliency scores, and corresponding spans. \textbf{Charades-STA} \cite{Charades-STA-dataset-2017} consists of 16,128 indoor video-query pairs with relevant moments. 
% For a fairer comparison, on Charades-STA, we use the official VGG features \footnote{https://prior.allenai.org/projects/charades} as RGB input.

For evaluation, we employ Recall@1 with thresholds of 0.5 and 0.7, mAP at IoU values of 0.5 and 0.75, and average mAP for MR. We use mAP and HIT@1 for HD, with HIT@1 assessing the accuracy rate of the highest-scoring moment.

\subsection{Implementation Details}
We set the attention layers in the multi-modal fusion module (MFM), cross-attention transformer (CAT), and transformer encoder to 2, 2, and 4, respectively. Following \cite{MomentDETR-2021}, we configure the number of span tokens $N_{s}$ as 10. Loss hyperparameters are set as follows: $\lambda_{cls} = 4$, $\lambda_{h} = 2$, $\lambda_{L1} = 10$, $\lambda_{IoU} = 1$, $w_{p} = 10$. We employ the AdamW \cite{AdamW-2017} optimizer, setting both the learning rate and weight decay at 1e-4. Our model undergoes training for 200 epochs, using a batch size of 32, on a single NVIDIA 3090 GPU. Since QVhighlights online testing \footnote{https://codalab.lisn.upsaclay.fr/competitions/6937} is limited to a maximum of five submissions, our ablation experiments are all conducted on QVhighlights \textit{val} split.

\begin{table}[t]
\caption{Performance comparison on QVHighlights \textit{test} split.}
\label{tab:QVHighlights-test}
\centering
\begin{adjustbox}{max width=\linewidth}
\begin{tabular}{@{}cccccccc@{}}
\toprule
\multicolumn{1}{c}{\multirow{3}{*}{\textbf{Methods}}} & \multicolumn{5}{c}{\textbf{MR}} & \multicolumn{2}{c}{\textbf{HD}} \\
\multicolumn{1}{c}{} & \multicolumn{2}{c}{R1} & \multicolumn{3}{c}{mAP} & \multicolumn{2}{c}{$\geq$ Very Good} \\
\cmidrule(lr){2-3}
\cmidrule(lr){4-6}
\cmidrule(lr){7-8}
\multicolumn{1}{c}{} & @0.5 & @0.7 & @0.5 & @0.75 & Avg. & mAP & HIT@1 \\
\midrule
% DVSE \cite{DVSE-2015} & - & - & - & - & - & 18.75 & 21.79 \\
% XML \cite{TVR-dataset-2020} & 41.83 & 30.35 & 44.63 & 31.73 & 32.14 & 34.49 & 55.25 \\ 
% Diwan et al. \cite{Zero-shot-VMR-2023} & 48.33 & 30.96 & 46.94 & 25.75 & 27.96 & - & - \\ 
% \midrule
Moment-DETR \cite{MomentDETR-2021} & 52.89 & 33.02 & 54.82 & 29.40 & 30.73 & 35.69 & 55.60 \\
SeViLA \cite{SeViLA-2023} & 54.50  & 36.50 & - & - & 32.30 & - & - \\
MomentDiff \cite{MomentDiff-2023} & 57.42 & 39.66 & 54.02 & 35.73 & 35.95 & - & - \\
UMT \cite{UMT-2022} & 56.23 & 41.18 & 53.83 & 37.01 & 36.12 & 38.18 & 59.99 \\
\textbf{MRNet} (Ours) & \textbf{61.54} & \textbf{45.20} & \textbf{61.21} & \textbf{39.89} & \textbf{39.53} & \textbf{39.23} & \textbf{63.45} \\ 
\bottomrule
\end{tabular}
\end{adjustbox}
\vspace{-3mm}
\end{table}

\begin{table}[t!]
\caption{Performance comparison on Charades-STA \textit{test} split.}
\label{tab:charades}
\centering
\begin{adjustbox}{max width=0.5\linewidth}
    \begin{tabular}{@{}lll@{}}
        \toprule
        Method & R1@0.5 & R1@0.7 \\ 
        \midrule
        2D-TAN \cite{2D-TAN-2020} & 39.81 & 23.31 \\
        UMT \cite{UMT-2022} & 49.35 & 26.16 \\
        Moment-DETR \cite{MomentDETR-2021} & 53.63 & 31.37 \\
        \textbf{MRNet} (Ours) & \textbf{55.84} & \textbf{33.59} \\ 
        \bottomrule
    \end{tabular}
\end{adjustbox}
\vspace{-2mm}
\end{table}

\vspace{-1mm}
\subsection{Comparison with State-of-the-arts}
\vspace{-1mm}
We report a comparison between MRNet and existing works on QVHighlights \textit{test} split, as detailed in Tab. \ref{tab:QVHighlights-test}. To ensure fairness, all models were trained from scratch. Results illustrate that  MRNet surpasses the SOTA method UMT \cite{UMT-2022} across all metrics. Notably, MRNet achieves significant improvements in MR-R1@0.5, MR-mAP@Avg, and HD($\geq$VG)-HIT@1, with increases of +5.31, +3.41, and +3.46, respectively. This verifies that MRNet can obtain a more accurate understanding of complex video and queries through our multi-modal fusion and query refinement. We also conduct experiments on the Charades-STA dataset, and as shown in Tab. \ref{tab:charades}, MRNet displays superior performance compared with other SOTA methods.

\vspace{-1mm}
\subsection{Ablation Studies}
\vspace{-1mm}
We conduct comprehensive ablations to investigate the significance of various modules and features in our model. Firstly, we remove transformer decoder from Moment-DETR \cite{MomentDETR-2021} and doubled transformer encoder layers, configuring this as our baseline model. As the upper part of Tab. \ref{tab:ablation-decoder} illustrates, this simple modification yielded modest gains, leading us to speculate that the decoder might be deficient in modeling temporal spans. 

\begin{table}[t!]
\caption{Effectiveness of each module in MRNet on QVHighlights \textit{val} split. VG is the abbreviation for very good.}
\label{tab:ablation-modules}
\centering
\begin{adjustbox}{max width=\linewidth}
\begin{tabular}{@{}cccccccc@{}}
\toprule
\multicolumn{3}{c}{\textbf{Modules}} & \multicolumn{3}{c}{\textbf{MR}} & \multicolumn{2}{c}{\textbf{HD} ($\geq$VG)} \\ 
\cmidrule(lr){1-3}
\cmidrule(lr){4-6}
\cmidrule(lr){7-8}
MFM & QRM & CAT & R1@0.5 & R1@0.7 & mAP Avg. & mAP & HIT@1 \\
\midrule
 &  &  & 55.79 & 37.84 & 33.28 & 35.96 & 56.35 \\
 &  & \ding{51} & 57.21 & 39.32 & 35.83 & 37.36 & 59.77 \\
 & \ding{51} & \ding{51} & 59.87 & 43.16 & 38.45 & 37.60 & 60.65 \\
\ding{51} &  & \ding{51} & 61.42 & 44.97 & 39.58 & 38.42 & 63.03 \\
\ding{51} & \ding{51} & \ding{51} & \textbf{62.00} & \textbf{47.68} & \textbf{40.34} & \textbf{39.58} & \textbf{64.03} \\
\bottomrule
\end{tabular}
\end{adjustbox}
\vspace{-4mm}
\end{table}

\begin{table}[t!]
\caption{Ablation study of different visual fusion modules on QVHighlights \textit{val} split.}
\label{tab:ablation-visual-encoder}
\centering
\begin{adjustbox}{max width=0.9\linewidth}
\begin{tabular}{@{}cccccc@{}}
\toprule
\multirow{2}{*}{\textbf{Fusion Modules}} & \multicolumn{3}{c}{\textbf{MR}} & \multicolumn{2}{c}{\textbf{HD} ($\geq$VG)} \\
\cmidrule(lr){2-4}
\cmidrule(lr){5-6}
 & R1@0.5 & R1@0.7 & mAP Avg. & mAP & HIT@1 \\
\midrule
CNN \cite{RASL-2019} & 58.39 & 39.10 & 34.33 & 36.45 & 59.23 \\
Co-attention \cite{DCTNet-2022} & 60.25 & 41.94 & 37.73 & 37.98 & 61.00 \\
\textbf{MFM} & \textbf{62.00} & \textbf{47.68} & \textbf{40.34} & \textbf{39.58} & \textbf{64.03} \\
\bottomrule
\end{tabular}
\end{adjustbox}
\vspace{-4mm}
\end{table}

Tab. \ref{tab:ablation-modules} verifies the necessity of each module, as our MRNet performs best when all modules are included. Row 1 obtains poor results by directly concatenating all multi-modal features and span tokens. Row 2 integrates CAT to underscore the significance of generating query-relevant cross-modal features in the visual-linguistic interaction module. Rows 3 and 4, adding MFM and QRM, respectively, demonstrate marked performance enhancement via dynamic multi-modal visual fusion and exploiting hierarchical features in text. Next, we compare various visual fusion modules, where MFM outperforms others, as seen in Tab. \ref{tab:ablation-visual-encoder}. This indicates that MFM is helpful for the model to learn complementary information from video. 

Subsequently, Tab. \ref{tab:ablation-features} evaluates different visual features. Rows 1-3 use single features and replace MFM with transformer encoder. Notably, only input RGB features yield superior results, indicating RGB in the video contains richer pixel-wise information. Rows 5 and 6 suggest that, compared to depth, optical flow has more outstanding assistance to the model, which is highly intuitive since the video contains numerous dynamic scenes. Row 4 interestingly notes that using both flow and depth underperforms RGB only. Row 7 shows optimal performance using all three features. Finally, adding a transformer decoder to MRNet slightly reduces performance on HD, as shown in rows 3 and 4 of Tab. \ref{tab:ablation-decoder}. This shows that the decoder will reduce the convergence on HD, but designing a better decoder may improve the effect of MR.

\begin{figure}[t!]
  \centering
  \includegraphics[width=\linewidth]{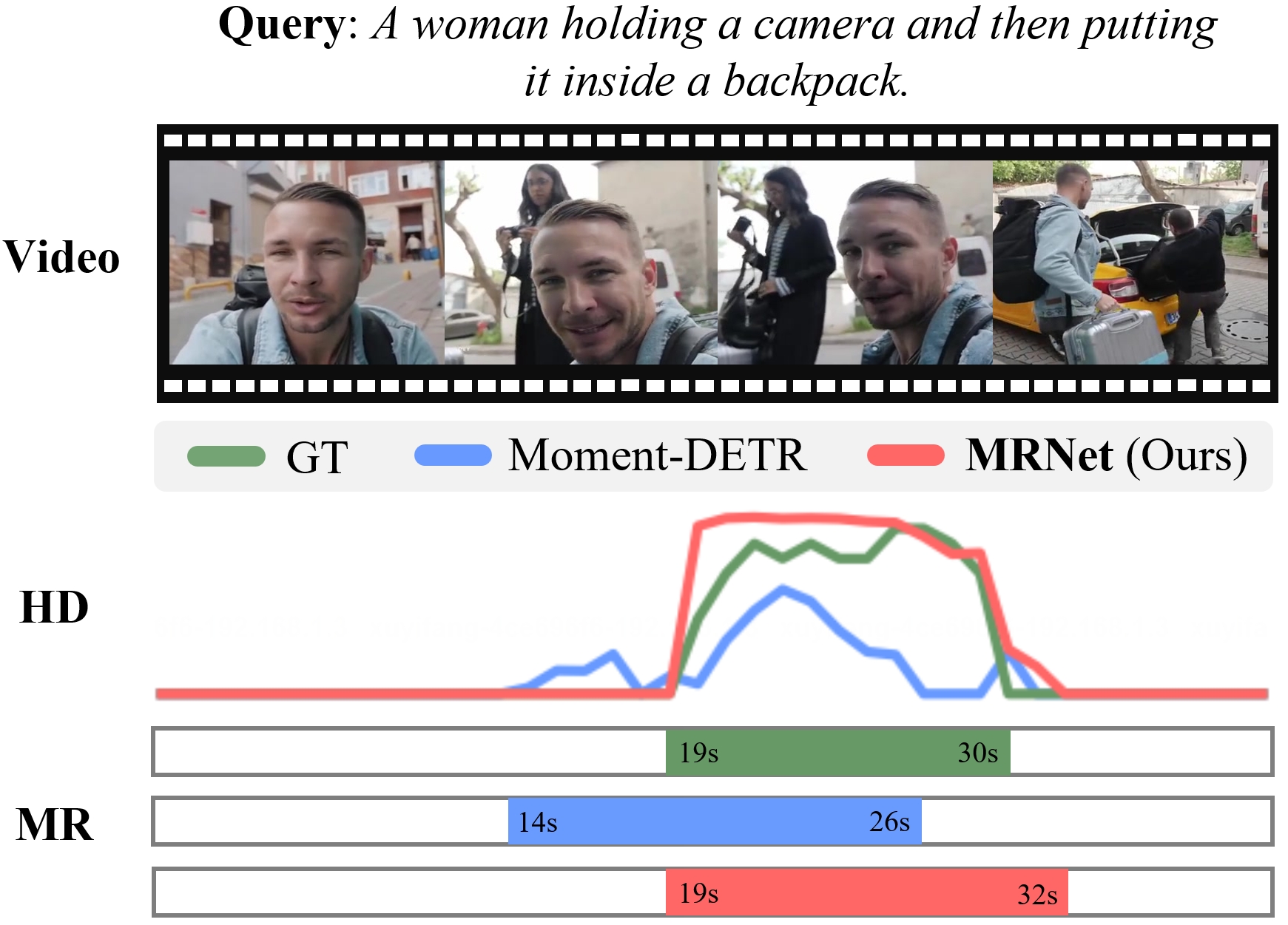}
  \vspace{-6mm}
  \caption{
    Qualitative comparison of the results on QVHighlights \textit{val} split.
  }
  \label{Fig:Fig5}
  \vspace{-3mm}
\end{figure}

To qualitatively validate the effectiveness of our method, we visualize a result of Moment-DETR and MRNet in Fig. \ref{Fig:Fig5}. We can observe that MRNet obtains more accurate highlights and spans than Moment-DETR. The main reason is that Moment-DETR only utilizes RGB, which fails to fully understand the static scenes and character activities throughout the video. Unlike Moment-DETR, MRNet learns complementary context from multi-modal cues to obtain more accurate predictions.

\begin{table}[t!]
\caption{Ablation study of different visual features on QVHighlights \textit{val} split.}
\label{tab:ablation-features}
\centering
\begin{adjustbox}{max width=\linewidth}
\begin{tabular}{@{}cccccccc@{}}
\toprule
\multicolumn{3}{c}{\textbf{Features}} & \multicolumn{3}{c}{\textbf{MR}} & \multicolumn{2}{c}{\textbf{HD} ($\geq$VG)} \\
\cmidrule(lr){1-3}
\cmidrule(lr){4-6}
\cmidrule(lr){7-8}
Depth & Flow & RGB & R1@0.5 & R1@0.7 & mAP Avg. & mAP & HIT@1 \\
\midrule
\ding{51} &  &  & 55.61 & 34.45 & 32.63 & 35.30 & 56.19 \\
 & \ding{51} &  & 57.94 & 38.39 & 35.15 & 37.11 & 59.81 \\
 &  & \ding{51} & 59.87 & 43.16 & 38.45 & 37.60 & 58.65 \\
\midrule
\ding{51} & \ding{51} &  & 57.21 & 39.98 & 35.89 & 36.67 & 58.41 \\
\ding{51} &  & \ding{51} & \textbf{63.10} & 46.00 & 39.87 & 38.88 & 62.77 \\
 & \ding{51} & \ding{51} & 60.84 & 46.45 & 40.10 & 38.72 & 63.16 \\
\ding{51} & \ding{51} & \ding{51} & 62.00 & \textbf{47.68} & \textbf{40.34} & \textbf{39.58} & \textbf{64.03} \\
\bottomrule
\end{tabular}
\end{adjustbox}
\vspace{-3mm}
\end{table}

\begin{table}[t!]
\caption{Effectiveness of transformer decoder on QVHighlights \textit{val} split.}
\label{tab:ablation-decoder}
\centering
\begin{adjustbox}{max width=\linewidth}
\begin{tabular}{@{}ccccccc@{}}
\toprule
\multirow{2}{*}{\textbf{Models}} & \multirow{2}{*}{\textbf{Decoder}} & \multicolumn{3}{c}{\textbf{MR}} & \multicolumn{2}{c}{\textbf{HD} ($\geq$VG)} \\
\cmidrule(l){3-5} 
\cmidrule(l){6-7} 
 &  & R1@0.5 & R1@0.7 & mAP Avg. & mAP & HIT@1 \\ \midrule
\multirow{2}{*}{Moment-DETR \cite{MomentDETR-2021}} & \ding{51} & 53.94 & 34.84 & 32.20 & 35.65 & 55.55 \\
 & \ding{55} & 54.19 & 34.19 & 32.26 & 36.59 & 57.68 \\ 
\midrule
\multirow{2}{*}{\textbf{MRNet} (Ours)} & \ding{51} & \textbf{62.69} & 46.40 & \textbf{40.42} & 38.48 & 62.79 \\
 & \ding{55} & 62.00 & \textbf{47.68} & 40.34 & \textbf{39.58} & \textbf{64.03} \\
\bottomrule
\end{tabular}
\end{adjustbox}
\end{table}

\vspace{-2mm}
\section{Conclusion and Future work}
\vspace{-2mm}
In this paper, we propose a novel MRNet for MR\&HD. Unlike previous works that only utilize RGB images as visual sources, the multi-modal fusion module in MRNet fuses multiple visual cues, including RGB images, optical flow, and depth maps, which empower the inference of dynamic scenes and the understanding of static scenes. In addition, we propose a query refinement module to aggregate semantic context in textual features at different granularities. Extensive experiments on the QVHighlights and Charades dataset demonstrate the superiority of our MRNet. In future work, we will employ large vision-language models to merge multi-modal inputs and design an efficient span-aware decoder.

\bibliographystyle{IEEEbib}
\bibliography{citations}

\end{document}